\documentclass[preprints,article,accept,pdftex,moreauthors]{MDPI/mdpi} 

\firstpage{1} 
\pubvolume{1}
\issuenum{1}
\articlenumber{0}
\pubyear{2024}
\copyrightyear{2024}
\datereceived{ } 
\daterevised{ } 
\dateaccepted{ } 
\datepublished{ } 
\hreflink{https://doi.org/} 

\usepackage{amsfonts}
\usepackage{algorithmic}
\usepackage{textcomp}
\usepackage{multicol}
\usepackage{rotating}

\def\figs{figures}

\newcommand{\figDataViz}{%
\begin{figure*}[ht]
    \centering
    \includegraphics[width=0.8\textwidth,trim=0 0 0 0,clip]{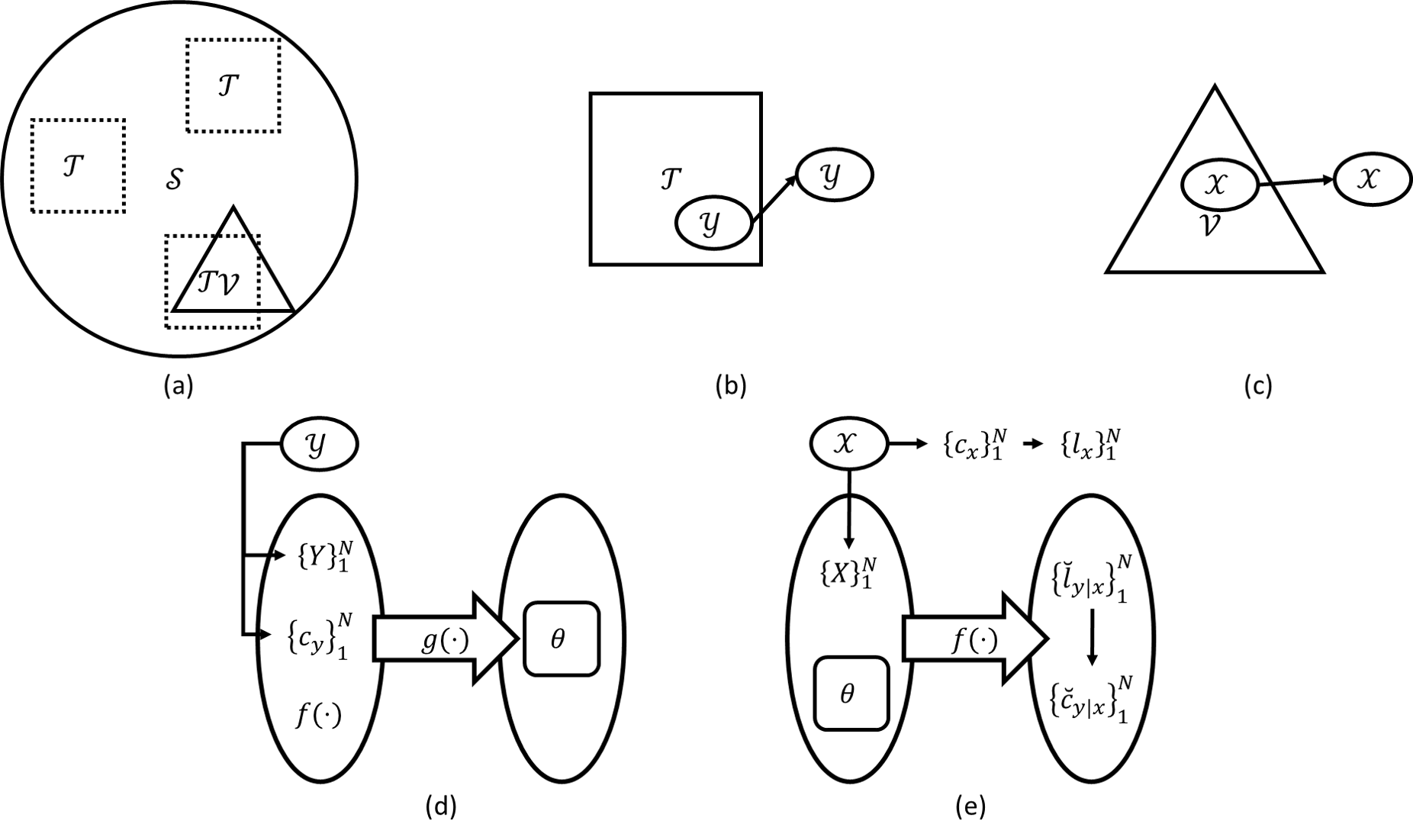}
    \caption{(\textbf{a}) A visualization of how the generalized problem space, $\mathcal{S}$, encompasses the application space, $\mathcal{V}$, as well as all possible data collection methods, $\mathcal{T}$. (\textbf{b}) The process of sampling from a collection method, $\mathcal{T}$, in order to produce a training dataset, $\mathcal{Y}$. (\textbf{c}) The sampling of data from the application space to produce an evaluation dataset, $\mathcal{X}$, for estimating a trained model's performance if used within the application space. (\textbf{d}) The training process, $g(\cdot)$, with a given architecture, $f(\cdot)$, and training set, $\mathcal{Y}$, to produce the parameters, $\theta$, that can be used for inference with the architecture, $f(\cdot;\theta)\equiv\phi(\cdot)$. (\textbf{e}) The inference process using a trained model on the evaluation dataset, $\breve{l}_{y|x}$.}
    \label{fig:dataviz}
\end{figure*}%
}

\newcommand{\figarch}{%
\begin{figure*}[ht]
    \includegraphics[width=\textwidth,trim=20 40 12 15,clip]{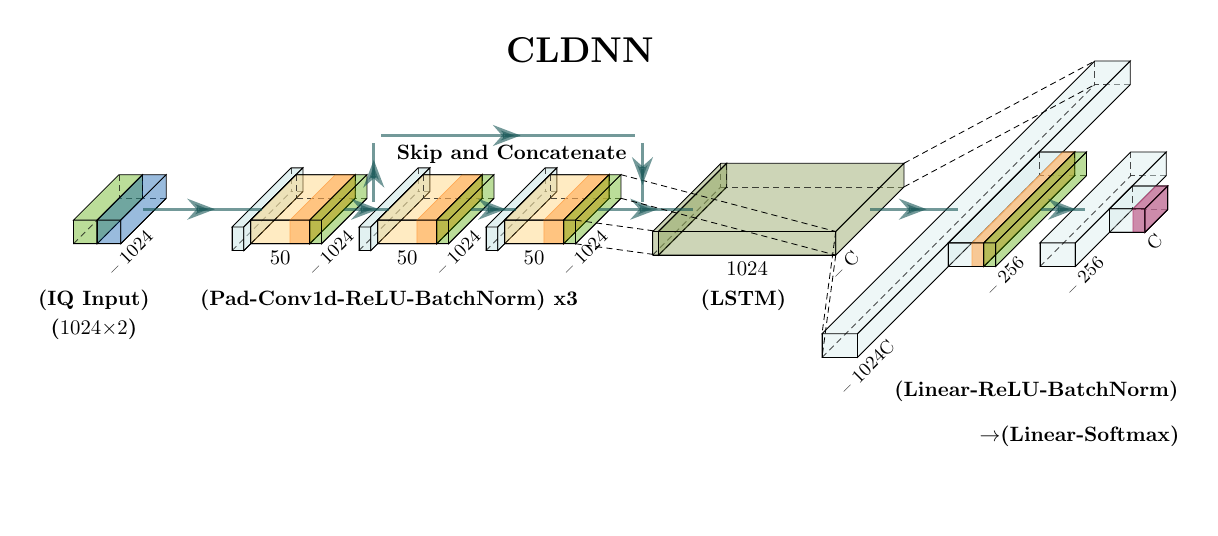}
    \caption{DL Neural Network Architecture for the Convolutional Long Short-Term Memory (LSTM) Deep Neural Network (CLDNN) Architecture used in this work as the DL approach for the 10-class waveform AMC problem space. The model consists of multiple layers of convolution layers each followed by non-linear activation and regularization that are combined along the channel dimension before passing through a recurrent LSTM layer and finally passing through linear layers followed by non-linear activation and regularization to produce the model's inference.}
    \label{fig:arch}
\end{figure*}%
}

\newcommand{\figAMrelationship}{%
\begin{figure*}[htbp]
    \centering
    \includegraphics[width=0.7\textwidth,trim=0 0 0 0,clip]{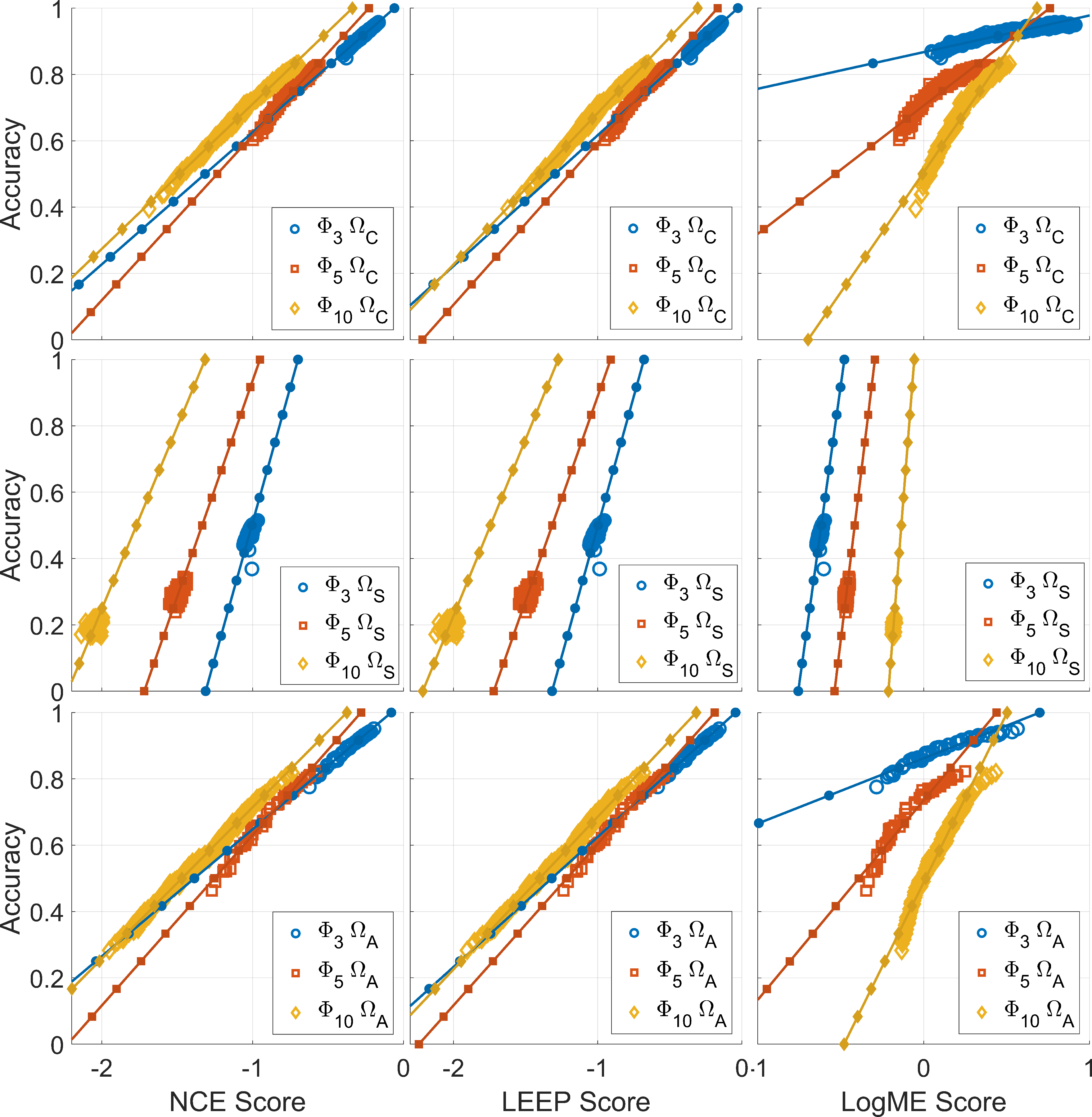}
    \caption{Visualization of the relationships between the three metrics (Left column: NCE, Middle column: LEEP, Right column: LogME) and the performance metric (Accuracy) of each network when measured on the results of the evaluation set $\Omega_{TC}$, or the \textit{target} dataset in TL vernacular. Each dataset used for training are positioned along the rows (Top row: $\Omega_{C}$, Middle row: $\Omega_{S}$, Bottom row: $\Omega_{A}$). Linear trends shown between the metrics and accuracy for better clarity in the relationships.}
    \label{fig:acc_metric_relationship}
\end{figure*}%
}

\newcommand{\figQArelationship}{%
\begin{figure}[ht]
    \centering
    \includegraphics[width=0.45\textwidth,trim=0 0 0 0,clip]{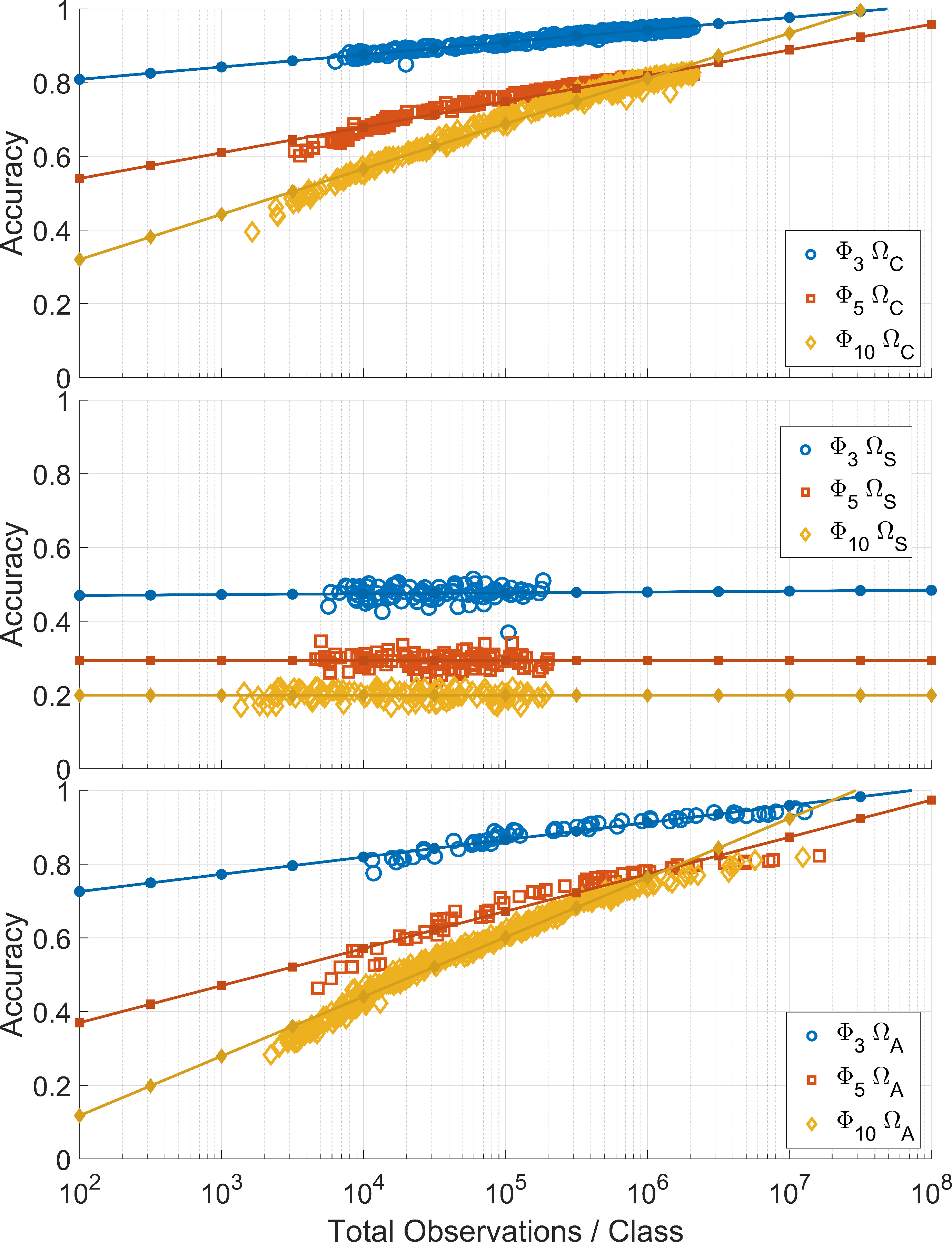}
    \caption{Plots show the relationship between quantity of data used from each dataset (Top: $\Omega_{C}$, Middle: $\Omega_{S}$, Bottom: $\Omega_{A}$) and the Accuracy achieved by networks trained on that amount of data. In general, the networks trained from datasets $\Omega_{C}$ and $\Omega_{A}$ have an increasing relation, but network, trained using $\Omega_{S}$ have a stagnant relation to performance in regards to quantity of data used to train.}
    \label{fig:qty_acc_relationship}
\end{figure}%
}

\newcommand{\figPTenAresiduals}{%
\begin{figure}[ht]
    \centering
    \includegraphics[width=0.45\textwidth,trim=0 0 0 0,clip]{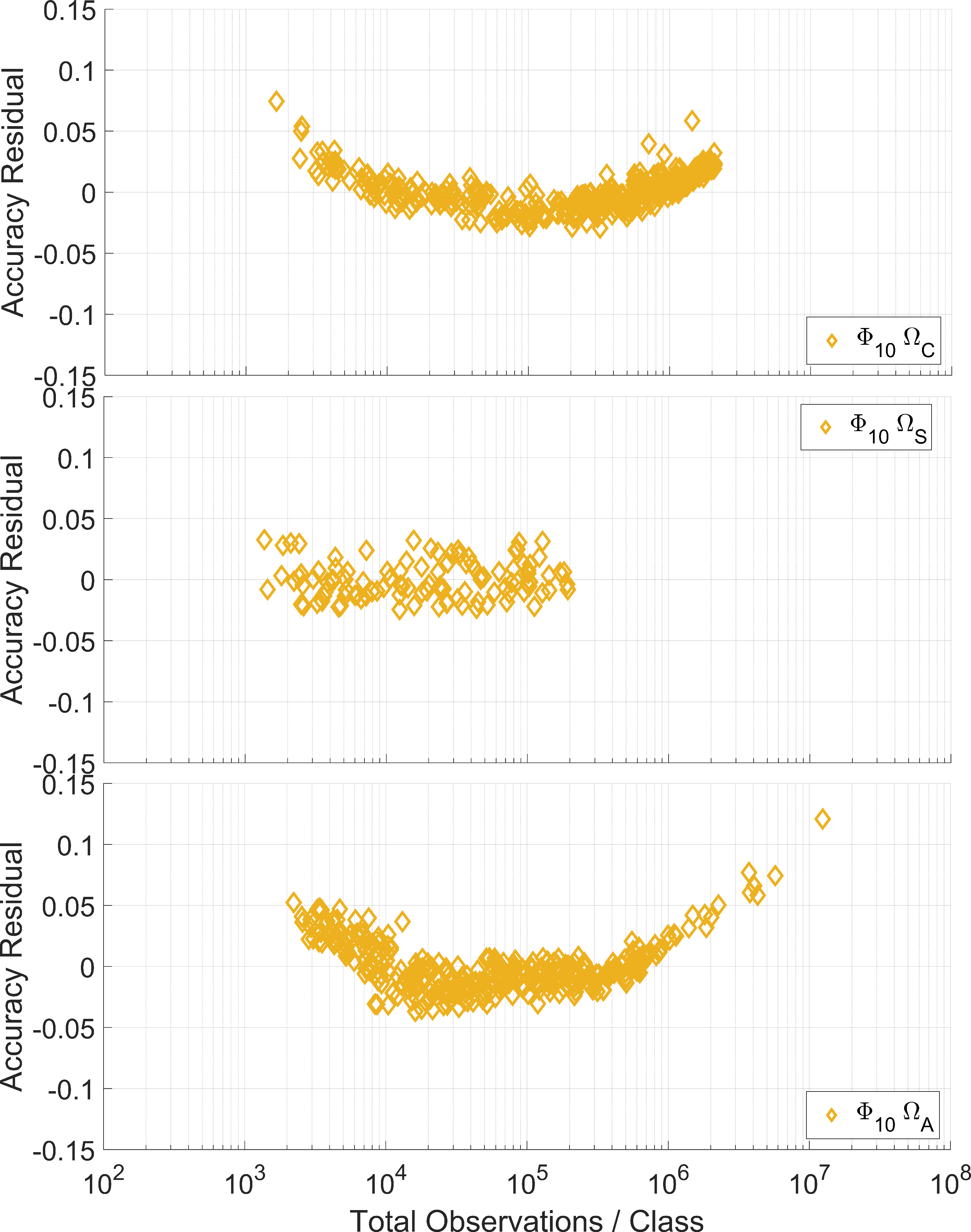}
    \caption{Plots show the residuals between the regressed log-linear fits of quantity of data available during training and the accuracy of each trained network and the observed accuracy of each network. Plots show similar trends across the used datasets (Top: $\Omega_{C}$, Middle: $\Omega_{S}$, Bottom: $\Omega_{A}$), with the edges of the available data deviating in the same direction, indicating a log-linear fit is not the ideal relationship between data quantity and accuracy.}
    \label{fig:qty_acc_residual}
\end{figure}%
}

\newcommand{\figWhitening}{%
\begin{figure*}[htbp]
    \centering
    \includegraphics[width=0.95\textwidth,trim=0 0 0 0,clip]{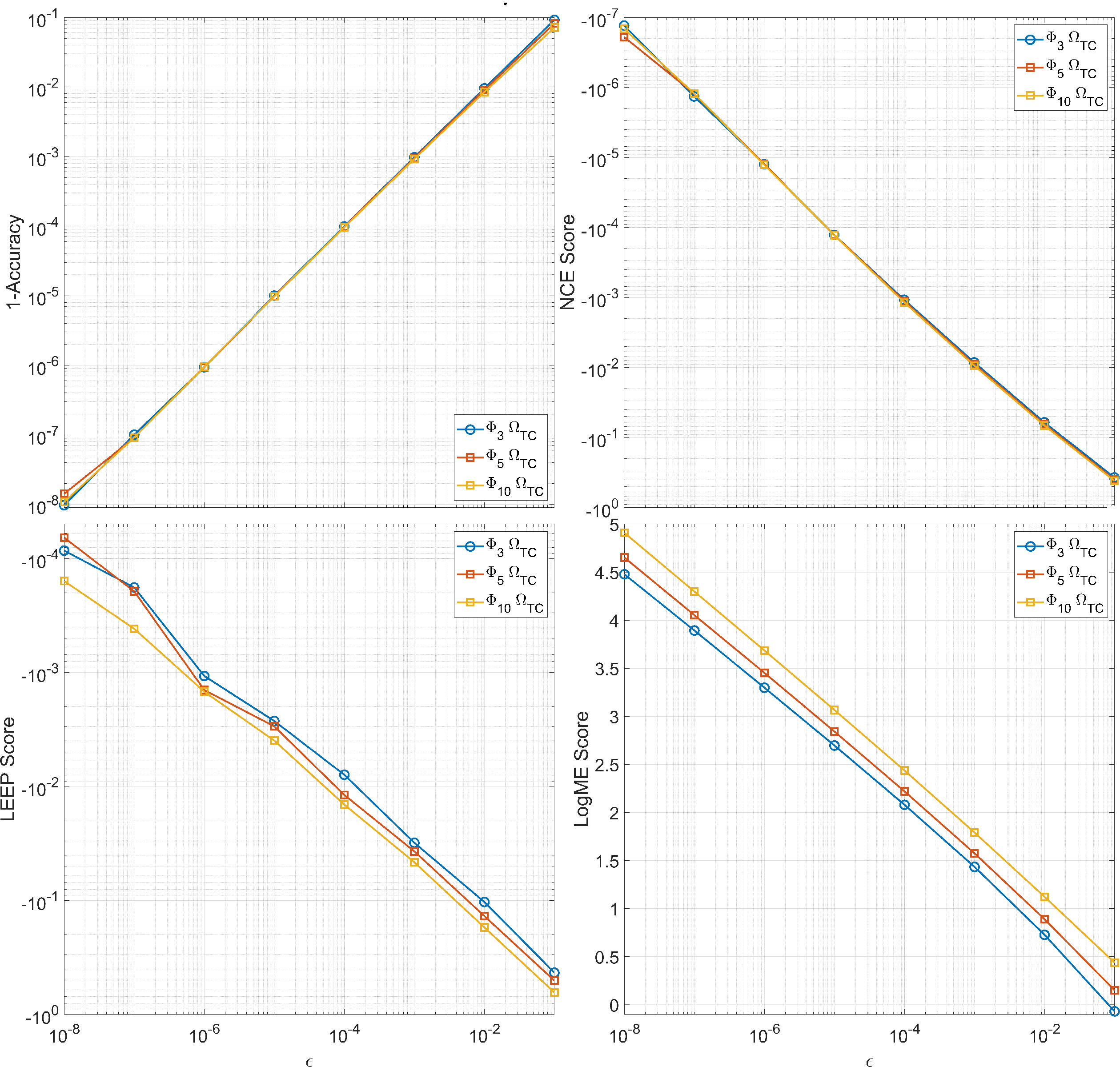}
    \caption{The change in the value of \{Top Left: Accuracy (1-Accuracy, logscale); Tope Right: NCE (logscale); Bottom Left: LEEP (logscale); Bottom Right: LogME (linear)\} as a function of the induced error, $\epsilon$, expected to accuracy from whitening the truth labels of the evaluation set $\Omega_{TC}$. Results plotted are the average values over 1000 iterations per data point.}
    \label{fig:whitening}
\end{figure*}%
}

\newcommand{\figMetricRelationships}{%
\begin{sidewaysfigure*}[htbp]
    \centering
    \includegraphics[width=0.99\textwidth,trim=0 0 0 0,clip]{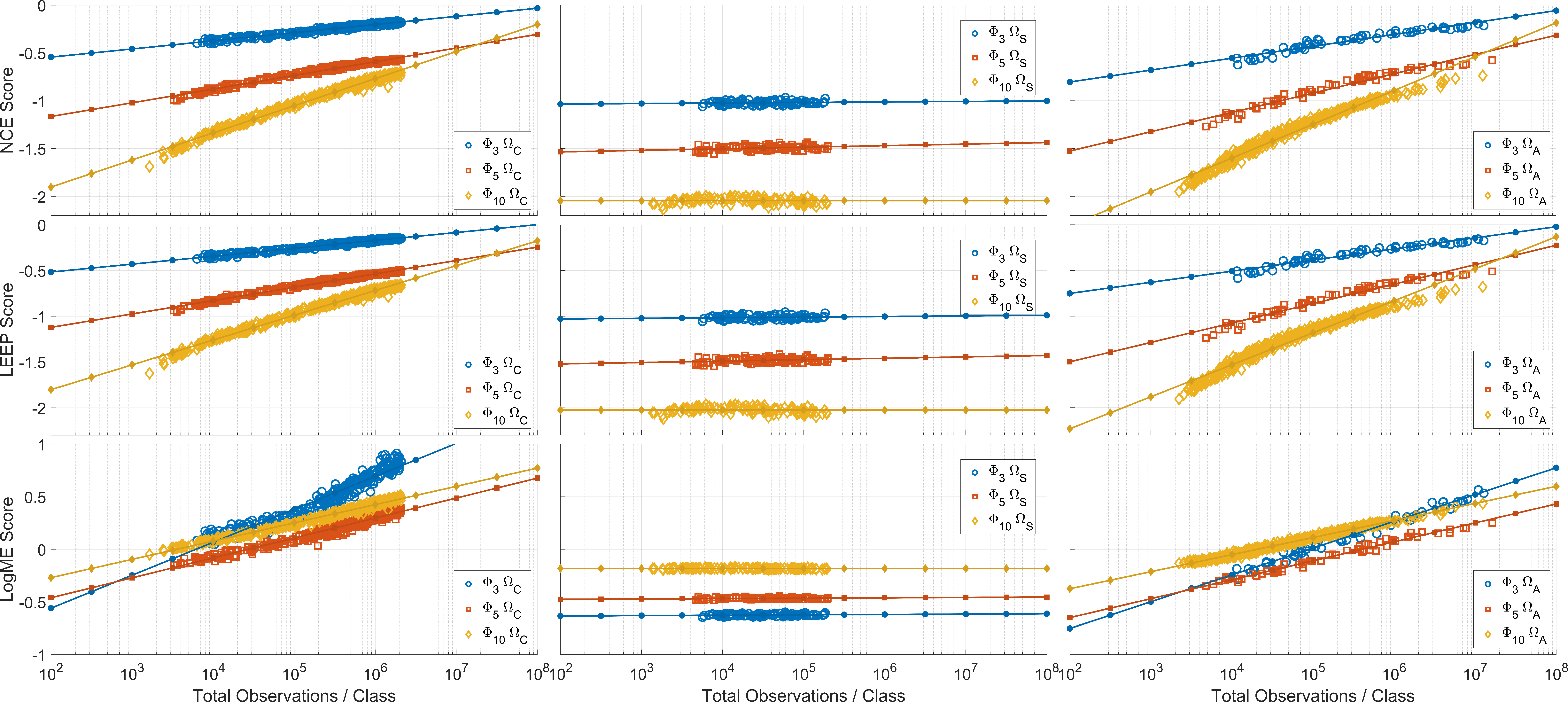}
    \caption{Plots show the relationship between quantity of data used from each dataset (Left: $\Omega_{C}$, Center: $\Omega_{S}$, Right: $\Omega_{A}$) and the metric (Top: NCE, Middle: LEEP, Bottom: LogME) achieved by networks trained on that amount of data. In general, the networks trained from datasets $\Omega_{C}$ and $\Omega_{A}$ have an increasing relation, but network, trained using $\Omega_{S}$ have a stagnant relation to performance in regards to quantity of data used to train. With a solution for a system that can perform arbitrarily close to perfect on the evaluation set, these linear regressions can then predict how much data would be required to achieve such a system.}
    \label{fig:qty_metric_relationship}
\end{sidewaysfigure*}%
}

\newcommand{\figMetricPredictions}{%
\begin{sidewaysfigure*}[htbp]
    \centering
    \includegraphics[width=0.99\textwidth,trim=0 0 0 0,clip]{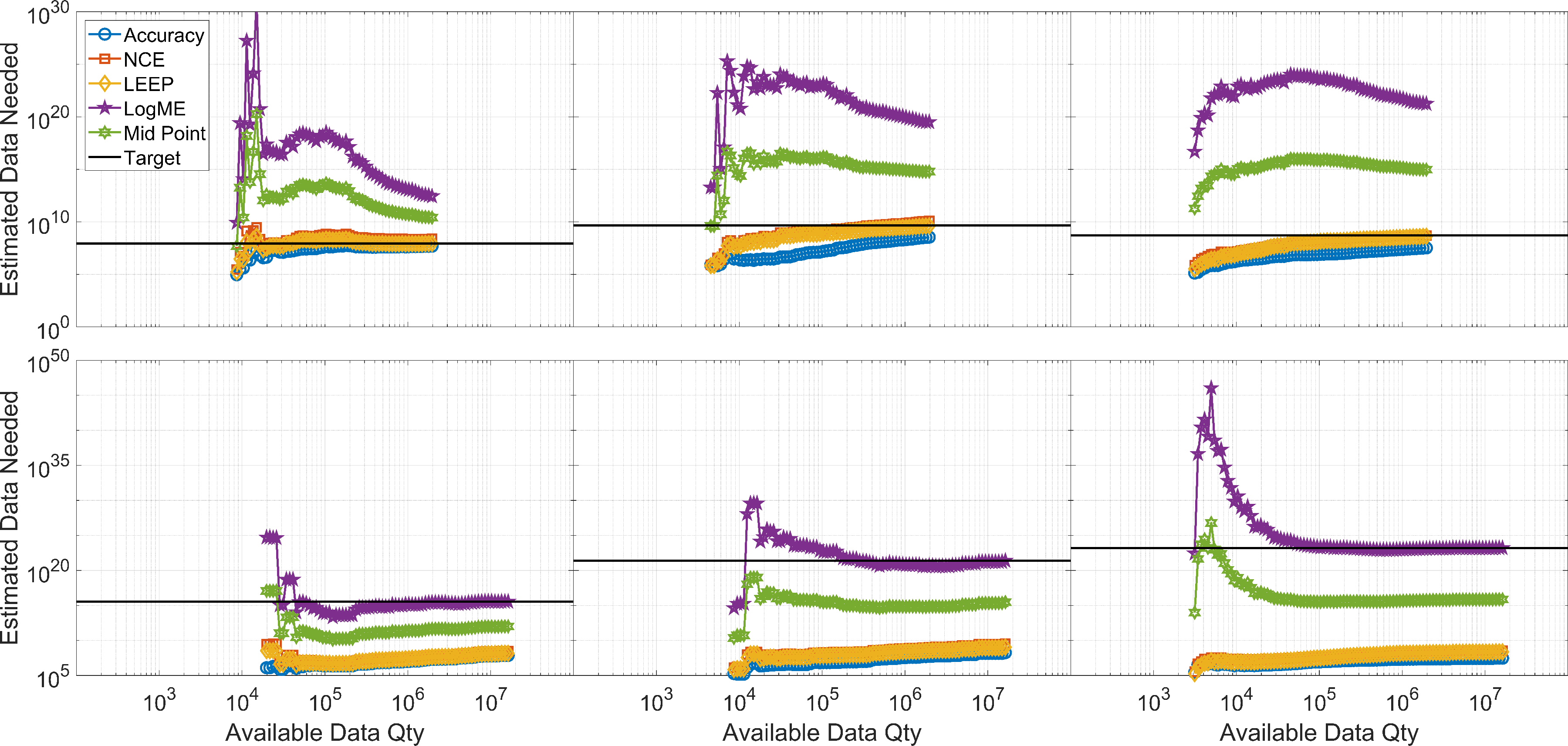}
    \caption{Plots show the quantity predictions based on a limited amount of available data used to regress the estimate on the \{Top Row: Capture $\Omega_C$; Bottom Row: Augmented $\Omega_A$\} datasets when being used to estimate across the \{Left: $\Phi_{3}$; Center: $\Phi_{5}$; Right: $\Phi_{10}$\} waveform space. The lines represent using \{Accuracy: circle, NCE: square, LEEP: diamond, LogME: pentagram, Log Scale Midpoint of NCE and LogME: hexagram, Target: none\} to predict data quantity needed, while Target is determined in Table \ref{tab:limited}. The midpoint serves as a balance between the underestimates produced by Accuracy/NCE/LEEP and overestimates from LogME for when a Target is not known \textit{a priori}.}
    \label{fig:qty_predictions}
\end{sidewaysfigure*}%
}

\newcommand{\tabdatasets}{%
\begin{table*}
  \small\sf\centering
  \caption{Description of datasets used within this work.}\label{tab:datasets}
  \begin{tabular}{p{0.08\textwidth} p{0.25\textwidth} p{0.5\textwidth}}
  \toprule
  \multicolumn{3}{l}{{\bf Training}}\\
  \midrule
  Symbol & Source & Description\\
  \midrule
  $\Omega_{C}$ & Capture & Consists of only capture examples\\
  \midrule
  $\Omega_{S}$ & Synthetic generation using KDE & Consists of simulated examples using the KDE of the capture dataset\\
  \midrule
  $\Omega_{A}$ & Augmentation using KDE & Consists of augmented examples from the capture dataset using the KDE\\
  \midrule
  \multicolumn{3}{l}{{\bf Evaluation}}\\
  \midrule
  $\Omega_{TC}$ & Capture & Consists of only capture examples $\Omega_{TC}\cap\Omega_{C}=\emptyset$\\
  \bottomrule
  \multicolumn{3}{l}{KDE, kernel density estimate}
  \end{tabular}
\end{table*}%
}

\newcommand{\tabwaveformsets}{%
\begin{table}
  \small\sf\centering
  \caption{Three waveform sets used in the work.}\label{tab:waveforms}
  \begin{tabular}{p{0.05\textwidth} p{0.35\textwidth}}
  \toprule
  Set & Waveforms\\
  \midrule
  $\Phi_{3}$ & BPSK, QPSK, Noise\\
  \midrule
  $\Phi_{5}$ & $\Phi_{3}$, QAM16, QAM64 \\
  \midrule
  $\Phi_{10}$ & $\Phi_{10}$, AM-DSB, BFSK, FM-NB, BGFSK, GMSK\\
  \bottomrule
  \end{tabular}
\end{table}%
}

\newcommand{\tabkendallstau}{%
\begin{table}
    \small\sf\centering
    \caption{Kendall's $\tau$ weighted correlation across datasets ($\Omega$) and waveform sets for Accuracy and (NCE, LEEP, LogME). Strong correlations will have an absolute value near 1, while no discernable correlation will be around 0. \textbf{Bold} values represent the combination of problem set and metric with the highest correlation with accuracy on the evaluation set.}\label{tab:wkt}
    \begin{tabular}{>{\centering}p{0.03\textwidth}>{\centering}p{0.05\textwidth}>{\centering}p{0.07\textwidth}>{\centering}p{0.07\textwidth}>{\centering\arraybackslash}p{0.07\textwidth}}
    \toprule
    Set & $\Omega$ & NCE& LEEP& LogME\\
    \midrule
    \multirow{3}{*}{$\Phi_{3}$} & $\Omega_{C}$ & \textbf{0.9774} & 0.9533 & 0.8033\\ 
    & $\Omega_{S}$ & \textbf{0.8249} & 0.8144 & 0.7382\\ 
    & $\Omega_{A}$ & \textbf{0.9666} & 0.9639 & 0.9377\\
    \midrule
    \multirow{3}{*}{$\Phi_{5}$} & $\Omega_{C}$ & 0.9438 & \textbf{0.9443} & 0.8788\\ 
    & $\Omega_{S}$ & \textbf{0.6554} & 0.6553 & 0.6334\\ 
    & $\Omega_{A}$ & \textbf{0.9794} & 0.9791 & 0.9582\\
    \midrule
    \multirow{3}{*}{$\Phi_{10}$} & $\Omega_{C}$ & \textbf{0.9794} & 0.9688 & 0.9609\\ 
    & $\Omega_{S}$ & 0.5165 & 0.4262 & \textbf{0.5298}\\ 
    & $\Omega_{A}$ & \textbf{0.9836} & 0.9808 & 0.9764\\
    \bottomrule
    \end{tabular}
\end{table}%
}

\newcommand{\tabgof}{%
\begin{table}
    \small\sf\centering
    \caption{Goodness-of-fit (GoF) for a log-linear regression between dataset quantity available for training across datasets ($\Omega$) and waveform sets ($\Phi$) for Accuracy ($\alpha$) and (NCE, LEEP, LogME). Perfect fit would have a value of 0. \textbf{Bold} values represent the best GoF value for the log-linear regression between the metric and data quantity available during training.}\label{tab:gof}
    \begin{tabular}{>{\centering}p{0.03\textwidth}>{\centering}p{0.03\textwidth}>{\centering}p{0.06\textwidth}>{\centering}p{0.06\textwidth}>{\centering}p{0.06\textwidth}>{\centering\arraybackslash}p{0.06\textwidth}}
    \toprule
    $\Phi$ & $\Omega$ & $\alpha$ & NCE & LEEP & LogME\\
    \midrule
    \multirow{3}{*}{$\Phi_{3}$} & $\Omega_{C}$ & 0.2478 & 0.2054 & \textbf{0.1885} & 0.2662\\ 
    & $\Omega_{S}$ & 1.0196 & \textbf{0.9515} & 0.9521 & 0.9531\\ 
    & $\Omega_{A}$ & 0.3120 & 0.2674 & 0.2636 & \textbf{0.1672}\\
    \midrule
    \multirow{3}{*}{$\Phi_{5}$} & $\Omega_{C}$ & 0.2499 & 0.1552 & \textbf{0.1458} & 0.1987\\ 
    & $\Omega_{S}$ & 0.9491 & \textbf{0.9367} & 0.9451 & 0.9652\\ 
    & $\Omega_{A}$ & 0.3016 & 0.2163 & 0.2208 & \textbf{0.1433}\\
    \midrule
    \multirow{3}{*}{$\Phi_{10}$} & $\Omega_{C}$ & 0.1514 & \textbf{0.1138} & 0.1173 & 0.1179\\ 
    & $\Omega_{S}$ & 0.9853 & 0.9783 & 0.9731 & \textbf{0.9697}\\ 
    & $\Omega_{A}$ & 0.2706 & 0.2652 & 0.2797 & \textbf{0.1102}\\
    \bottomrule
    \end{tabular}
\end{table}%
}

\newcommand{\tabperf}{%
\begin{table*}
    \small\sf\centering
    \caption{The performance of Accuracy ($\alpha$), NCE, LEEP, and LogME for the label whitening proceedure proposed in Section \ref{sec:data_qs} for a desired error $\epsilon = 1e-5$. Additionally the data quantity needed per metric for the log-linear regression of the metric with the data quantity used during training is provided.}\label{tab:perf}
    \begin{tabular}{>{\centering}p{0.03\textwidth}>{\centering}p{0.08\textwidth}>{\centering}p{0.09\textwidth}>{\centering}p{0.09\textwidth}>{\centering}p{0.06\textwidth}>{\centering}p{0.03\textwidth}>{\centering}p{0.09\textwidth}>{\centering}p{0.09\textwidth}>{\centering}p{0.09\textwidth}>{\centering\arraybackslash}p{0.09\textwidth}}
    \toprule
    \multirow{2}{*}{$\Phi$} & $\alpha$ & NCE & LEEP & LogME & \multirow{2}{*}{$\Omega$} & $\alpha$ & NCE & LEEP & LogME\\
    & Metric & Metric & Metric & Metric & & Quantity & Quantity & Quantity & Quantity\\
    \midrule
    \multirow{3}{*}{$\Phi_{3}$} & \multirow{3}{*}{0.999990} & \multirow{3}{*}{-1.275e-4} & \multirow{3}{*}{-2.668e-3} & \multirow{3}{*}{2.696} & $\Omega_{C}$ & 48.7e6 & 237e6 & 88.5e6 & 2.46e12\\ 
    & & & & & $\Omega_{S}$ & 6.42e224 & 1.64e190 & 1.38e158 & $\infty$\\ 
    & & & & & $\Omega_{A}$ & 72.8e6 & 291e6 & 144e6 & 3.47e15\\
    \midrule
    \multirow{3}{*}{$\Phi_{5}$} & \multirow{3}{*}{0.999990} & \multirow{3}{*}{-1.272e-4} & \multirow{3}{*}{-2.972e-3} & \multirow{3}{*}{2.842} & $\Omega_{C}$ & 394e6 & 13.5e9 & 4.45e9 & 2.57e19\\ 
    & & & & & $\Omega_{S}$ & $\infty$ & 9.15e95 & 6.57e100 & $\infty$\\ 
    & & & & & $\Omega_{A}$ & 181e6 & 3.60e9 & 1.11e9 & 2.41e21\\
    \midrule
    \multirow{3}{*}{$\Phi_{10}$} & \multirow{3}{*}{0.999990} & \multirow{3}{*}{-1.289e-4} & \multirow{3}{*}{-3.963e-3} & \multirow{3}{*}{3.066} & $\Omega_{C}$ & 34.2e6 & 514e6 & 430e6 & 1.61e21\\ 
    & & & & & $\Omega_{S}$ & $\infty$ & $\infty$ & $\infty$ & $\infty$\\ 
    & & & & & $\Omega_{A}$ & 29.3e6 & 337e6 & 233e6 & 1.57e23\\
    \bottomrule
    \end{tabular}
\end{table*}%
}
\newcommand{\tablimited}{%
\begin{table}
    \small\sf\centering
    \caption{Quantity estimates being taken as truth for the combinations of waveform groups, $\Phi$, and training datasets, $\Omega$. The augmented quantities are estimated using the LogME metric regression, while the captured quantities are estimated from either the NCE or LEEP metric based on the GoF in Table \ref{tab:gof}.}\label{tab:limited}
    \begin{tabular}{ccc}
        \toprule
        $\Phi$ & $\Omega_C$ & $\Omega_A$ \\ \midrule
        $\Phi_{3}$ & 88.5e6 & 3.47e15 \\
        $\Phi_{5}$ & 4.45e9 & 2.41e21 \\
        $\Phi_{10}$ & 514e6 & 1.57e23 \\
        \bottomrule
    \end{tabular}
\end{table}%
}

\newcommand{\workingtitle}{Training from Zero: Radio Frequency Machine Learning Data Quantity Forecasting}
\Title{\workingtitle}

\TitleCitation{\workingtitle}

\Author{{William H. Clark, IV} $^{1}$\orcidA{}, {Alan J. Michaels} $^{1}$}

\AuthorNames{{William H. Clark, IV} and {Alan J. Michaels}}

\AuthorCitation{{Clark, IV, W.H.}; {Michaels, A.}}

\address[1]{
$^{1}$ \quad Virginia Tech National Security Institute, Blacksburg}

\corres{Correspondence: bill.clark@vt.edu}

\newcommand{\ABSTRACTstart}[1]{\abstract{#1}}
\newcommand{\ABSTRACTend}[1][]{#1}
\newcommand{\KEYWORDSstart}[1]{\keyword{#1}}
\newcommand{\KEWWORDSend}[1][]{#1}
\ABSTRACTstart{
The data used during training in any given application space is directly tied to the performance of the system once deployed.
While there are many other factors that go into producing high performance models within machine learning, there is no doubt that the data used to train a system provides the foundation from which to build.
One of the underlying rule of thumb heuristics used within the machine learning space is that more data leads to better models, but there is no easy answer for the question, ``How much data is needed?''
This work examines a modulation classification problem in the Radio Frequency domain space, attempting to answer the question of how much training data is required to achieve a desired level of performance, but the procedure readily applies to classification problems across modalities.
The ultimate goal is determining an approach that requires the least amount of data collection to better inform a more thorough collection effort to achieve the desired performance metric.
While this approach will require an initial dataset that is germane to the problem space to act as a \textit{target} dataset on which metrics are extracted, the goal is to allow for the initial data to be orders of magnitude smaller than what is required for delivering a system that achieves the desired performance.
An additional benefit of the techniques presented here is that the quality of different datasets can be numerically evaluated and tied together with the quantity of data, and ultimately, the performance of the architecture in the problem domain.
}\ABSTRACTend

\KEYWORDSstart{
data analysis, data collection, machine learning, neural networks, pattern recognition, physical layer, {RF} signals, {RFML}, signal synthesis, software radio, wireless communication}
\KEWWORDSend
\begin{document}




\newcommand{\specialfirstsection}[1]{#1}
\newcommand{\specialfirstword}[2]{#1#2}
\specialfirstsection{\section{Introduction}\label{sec:introduction}}
\specialfirstword{M}{achine} Learning (ML) is ``the capacity of computers to learn and adapt without following explicit instructions, by using algorithms and statistical models to analyse and infer from patterns in data'' \cite{oxford_dict_ml}.
No matter the field, ML begins and ends on the data available to use during training.
Without relevant data to learn from, ML is effectively a ``garbage in, garbage out'' system \cite{mlgigo}.
The application of ML to problems within the Radio Frequency (RF) domain is no exception to this rule, yet within the scope of intentional man-made emissions, data is easier to synthesize than within more prolific domains such as image processing \cite{oshea_synth}.
Due to the readily available tools for developing ML-based algorithms (Tensorflow \cite{tensorflow}, PyTorch \cite{pytorch}, etc.), and this ease of synthesis for establishing comprehensive datasets, there has been an explosion of published work in the field.
Adding to the ease of training models, the RF domain has the availability of open source toolsets for synthesizing RF waveforms such as GNU Radio \cite{gnuradio} and Liquid-DSP \cite{liquiddsp} to name a few.
However, going from a purely synthetic environment to a functional application running in the real world has a number of considerations that must be addressed.
A brief explanation about the gaps from synthetic data to functional application data is discussed in Section \ref{sec:rf}.

This paper focuses on providing an applied understanding of working with ML in the RF spectrum, with particular regard to understanding the training data in applications that fall in the domain space of Radio Frequency Machine Learning (RFML).
To better clarify this, RFML is a subset of ML that overlaps communications, radar, or any other application space that utilizes the RF spectrum in a statistically repeatable manner, where ML algorithms are applied as intelligently close to the digitized samples of the RF spectrum, or Physical Layer in the Open Systems Interconnection (OSI) model, as possible \cite{wong2020rfml}.
For simplicity within this work, focus will be given to the field of Deep Learning (DL) as the particular subset within ML due to the well suited nature of DL systems at extracting inference from raw data \cite{Goodfellow-et-al-2016}.
Two fundamental questions for developing a DL architecture for a given application are:
\begin{itemize}
    \item What data should be, or is available to be used in order to train the system;
    \item How much data is needed for the current approach to achieve the desired performance level?
\end{itemize}
These two questions are systematically addressed within this work, providing a blueprint for how they can be answered in general.

The two most common problems in regard to datasets with DL systems are having a large enough quantity of data and having that data be at a high enough quality, in order to develop a well generalized model \cite{Goodfellow-et-al-2016}.
In fact, answering the question of how much data is needed is a fundamental unknown that relies on the developers' experience and insight into the problem space.
The end result of attempts at answering this question are the common rule of thumb answers of `ten times the feature space' or `as much data as is available' and is an issue across all domains \cite{10386440, 9961269, s22186982, NG2022100198, 10.1145/3406325.3451036, 9019687, 9154298, wang2024data, mühlenstädt2024data, 9880008}.

The work of Chen \textit{et al} \cite{ml_quality_1} describe the concept of quality in three ways: \textit{Comprehensiveness, Correctness}, and \textit{Variety}.
In this work the primary focus is on understanding the aspect of \textit{Variety} in terms of the origin of the data, \textit{Captured} (received spectral samples by a sensor), \textit{Synthetic} (samples generated from formulae), and \textit{Augmented} (a mixture of the previous origins), while the other two aspects are more concerned with the information being both present in the dataset as well as being correctly labeled, which is given for the datasets used.
More details are discussed about the origin of the data are presented in Section \ref{sec:data_origin}, while data quantity and quality are discussed in more detail in Section \ref{sec:data_qs}.

The discussion of characteristics inherited in the application of ML are given in Section \ref{sec:ml}, with the RFML problem of Automatic Modulation Classification (AMC), which is discussed in more detail in Section \ref{sec:amc}.
The more well discussed problems and nuisances inherent to the RF spectrum are discussed in Section \ref{sec:rf}.
Examining the effects of data quantity, along with the concept of how data quality can be quantified are presented in Section \ref{sec:data_qs}.
The problem of estimating the data needs from a minimal set are contrasted to the full availability of data in Section \ref{sec:application} where a combination of two metrics offers a more balanced estimate than either metric alone.
Finally, conclusions about how the presented AMC approaches can be well generalized to RFML at large are presented in Section \ref{sec:conclusion}.


\figDataViz

\subsection{Machine Learning Concepts}
\label{sec:ml}
\noindent ML, in the most general form, is the process of creating a function that maps an observable in the form raw data, meta data, and/or extracted features of the data to some more convenient form for an applied task through observations available during training \cite{Goodfellow-et-al-2016}.
Therefore, given a training dataset, $\{Y\}_1^N\sim \mathcal{Y}$, drawn from an observation space, $\mathcal{T}$, inherent to the generalized problem space, $\mathcal{T}\subset \mathcal{S}$, ML works to create a mapping $f : \mathcal{Y}\mapsto \mathcal{T}$.
The ultimate goal is that the learned mapping is well generalized and can also be applied to the general space $\mathcal{S}$ with the same performance as within the $\mathcal{T}$.
A visualization of the relationship between the generalized problem space, $\mathcal{S}$, and the observation space from which training data is collected, $\mathcal{T}$, is shown in Fig. \ref{fig:dataviz} (a) with the training dataset, $\mathcal{Y}$, shown in Fig. \ref{fig:dataviz} (b).


\figarch

The mapping is learned through a training procedure, $g: f,\mathcal{Y}\mapsto \theta$, which produces the parameter space that defines the behavior of $f$, and is visualized in Fig. \ref{fig:dataviz} (d).
Given the focus on DL systems, $\theta$ are a set of weights and biases that are used within the DL architecture, $f$.
The DL architecture used in this work is shown in Fig. \ref{fig:arch} and whose architecture was shown to be well suited to the AMC problem space in the work of West and O'Shea \cite{west-amc2}, while the regularization was incorporated from the work of Flowers and Headley for the increased convergence rate \cite{rfml2019tutorial}.
The challenges and discussion within AMC are discussed in further detail in Section \ref{sec:rf}.

A brief overview of concepts utilized in this work with regard to ML are given below.
In particular, the concepts of DL and Transfer Learning (TL) are fundamental to the setup and analysis of the crux of this work.

\subsubsection{Deep Learning}
\noindent Goodfellow \textit{et al} \cite{Goodfellow-et-al-2016} goes into great detail for developing an understanding of the complexities surrounding DL.
A brief explanation follows that DL is a subset of ML that makes use of multiple layers of processing that can in theory approach the problem with simpler computations, which when accumulated, allow for solving a complex problem \cite{Goodfellow-et-al-2016}.
The DL approach used in this work uses Deep Neural Networks (DNN), which make use of multiple sequential layers consisting of convolutional layers, a recurrent Long Short-Term Memory (LSTM) layer whose hidden size is tied to the dimension of the classification space, and two successive linear layers before producing the model output.
After each convolutional and linear layer that is not the final layer of the network, a Rectified Linear Unit (ReLU) non-linear activation is used followed by Batch Normalization regularization layer.
The final layer is followed by a Softmax non-linear activation to give the output as an estimate of the probability that the current observation is one of the classes specified during training.
A visualization of this architecture is shown in Fig. \ref{fig:arch} for the 10-class AMC classification problem.
For simplicity, the application of the function and weights on an observation is shortened to, $\breve{l} = \phi(x) \equiv f(x;\mathbf{\theta}) \in \mathbb{R}^C$ where $C$ is the number of classes in the classification problem.

\subsubsection{Transfer Learning}\label{sec:tl}
\noindent TL is the practice of training a model on one dataset/ domain (\textit{source}), or otherwise taking a pre-trained model, and training with a new dataset/ domain (\textit{target}) instead of starting from a random initialization \cite{xfer_survey_01}.
Depending on assumptions between the \textit{source} and \textit{target}, the TL application can be categorized as \textit{homogeneous}, where differences exist in the distributions between \textit{source} and \textit{target}, or \textit{heterogeneous}, where the differences are in the feature space of the problem \cite{xfer_survey_01,xfer_survey_02}.
A valuable discussion for understanding the concepts of \textit{homogeneous} and \textit{heterogeneous} is provided by Wong and Michaels \cite{wong_tl} by explaining the change in distributions as a change of the dataset's collected/generated domain, while the feature space of the problem can be associated with the intended task the \textit{source} model is trained on and can contrasted to the task of the \textit{target} problem.
The two most common types of TL include retraining the classification head where early layers are frozen during training preserving feature extraction, or fine-tuning of the whole model \cite{leep}.
In this work, TL is applied in a \textit{homogeneous} problem space where the underlying distributions on the data vary, but the generalized problem space is the same between datasets, otherwise called a \textit{Domain Adaptation} from \cite{wong_tl} and more specifically an \textit{Environment Platform Co-Adaptation}.
Additionally, wherever the retraining is done in this work, the fine-tuning approach is utilized, allowing for adjustments to the feature space, which might not be observable in the \textit{source} dataset.
An important note here: this work does not evaluate any aspect of TL on the problem space, rather makes use of metrics developed for the purpose of TL.
Understanding how TL is best used within RFML is beyond the scope of this work.

The study of TL is complex and well explored \cite{xfer_survey_01,xfer_survey_02,nce,leep,logme,wong_tl}, and from the effort to understand how to choose an optimal pre-trained model for a desired application, the metrics Negative Conditional Entropy (NCE) \cite{nce}, Log Expected Empirical Prediction (LEEP) \cite{leep}, and Logarithm of Maximum Evidence (LogME) \cite{logme} are repurposed to analyze the relationship between available data quantity during training and system performance for a given evaluation set.

In this paper, the evaluation set has the same labels as the training set, but the distributions are not assumed to be equivalent.
Therefore the evaluation set, $\{X\}_1^N\sim \mathcal{X}$, is drawn from an observation space, $\mathcal{V}$, which is inherent to the generalized problem space, $\mathcal{V}\subset \mathcal{S}$, and is visualized in Fig. \ref{fig:dataviz} (a) and (c).
For clarity going forward, due to the shared labels between \textit{source} and \textit{target} in this work, the \textit{source} labels are found through a forward pass of the evaluation set through the network, therefore the $i^\text{th}$ observation's \textit{source} label is given by $\breve{l}_{y|xi} = \phi(x_i) \in \mathbb{R}^C$, with the inference given as $\breve{c}_{y|xi} = \text{argmax}(\breve{l}_{y|xi}) \in \mathbb{Z}^1$.
The process of extracting the evaluation inference of a trained model is visualized in Fig. \ref{fig:dataviz} (e).
By contrast, the \textit{target} label directly gives $c_{x,i} \in \mathbb{Z}^1$ by the truth of the $i^\text{th}$ observation and can be one-hot encoded to provide $l_{xi} = \text{OH}(c_{xi},C) \in \mathbb{R}^C$.
Given the above notation, NCE is given as
\begin{equation}\label{eq:nce}
\begin{split}
    \text{NCE}&(\breve{c}_{y|x},c_x) = \sum_{j=1}^{C}\hat{P}(\breve{c}_{y|x}=j)\\
    &\cdot\sum_{k=1}^{C}\hat{P}(c_x=k|\breve{c}_{y|x}=j)\log(\hat{P}(c_x=k|\breve{c}_{y|x}=j)),
\end{split}
\end{equation}
where $\hat{P}(\cdot)$ are the empirical distributions found as
\begin{equation}
    \hat{P}(\breve{c}_{y|x}=j) = \frac{1}{N}\sum_{i=1}^{N} \breve{c}_{y|xi} == j,
\end{equation}
\begin{equation}
    \hat{P}(c_x=k | \breve{c}_{y|x}=j) = \frac{1}{N}\sum_{i=1}^{N} (\breve{c}_{y|xi} == j)\cdot(c_{xi} == k).
\end{equation}
The \textit{source} labels are iterated over with $j$, while $k$ iterates over the \textit{target} labels.
LEEP is given as 
\begin{equation}
    \text{LEEP}(\breve{l}_{y|x},c_x) = \frac{1}{N}\sum_{i=1}^{N}\log\left(\breve{l}_{y|xi}\cdot\hat{P}(c_x=k|\breve{l}_{y|x})\right),
\end{equation}
where the empirical conditional probability, $\hat{P}(c_x=k|\breve{l}_{y|x})$ is given as
\begin{equation}
\begin{split}
    \hat{P}(c_x=k|\breve{l}_{y|x}) &= \left[\hat{P}(k|j=1),\ldots,\hat{P}(k|j=C)\right]^T\\
    \hat{P}(k|j) &= \hat{P}(k,j) / \sum_{k'=1}^{C} \hat{P}(k',j)\\
    \hat{P}(k,j) &= \frac{1}{N}\sum_{i=1}^N \breve{l}_{y|xi}[j]\cdot(c_{xi}==k).
\end{split}
\end{equation}
The LEEP score, for the combination of the model and evaluation set, is given as the average log of all probabilities of getting the correct label in the evaluation set given the empirical probability of the labels provided by the model under test \cite{leep}.
Whereas LogME is given as
\begin{equation}
    \text{LogME}(\breve{l}_{y|x},c_x) = \frac{1}{NC}\sum_{k=1}^{C}\log(p(c_{x}=k|\breve{l}_{y|x},\alpha,\beta)),
\end{equation}
where $\alpha$ and $\beta$ are iteratively solved to maximize the evidence, $p(c_x=k|\breve{l}_{y|x})$, for a linear transform applied to $\breve{l}_{y|x}$, which is then averaged over the number of classes, $C$, and normalized by the number of observations, $N$, in the evaluation set \cite{logme}.

In the most general sense, the importance of these metrics is how well correlated, either positively or negatively, the metric is with the desired performance of the network after being retrained on the \text{target} dataset.
Within this work the explanation provided by You \textit{et al} \cite{logme} for using Kendall's $\tau$ coefficient \cite{kendall_tau} is utilized as the most significant relationship between performance and the metric of choice is a shared general monotonicity that allows for a trend in the metric to indicate a trend in performance as well.

\subsubsection{Automatic Modulation Classification}\label{sec:amc}

\noindent The RF problem presented in this work is then the classification of the modulation present in the original transmitted waveform $s_{BB}(t)$.
AMC is then the problem of being able to identify how information (or lack there of) is being applied to a specific time and frequency slice of the overall spectrum.
When coupled with the problem of signal detection, is a waveform present or not in a time and frequency slice, the problem space is often referred to as Automatic Modulation Recognition; however, in this work the trained network is determining what is there, rather than the additional task of where is it, so AMC is a better category for the task.
While AMC is one of the oldest disciplines within RFML, traditional approaches relied on expert analysis and feature extraction \cite{dandawate_et_al,swami_sadler,headley_daSilva,dobre_et_al}, though over the last three decades heavier reliance on ML has been used for feature fusion and decision making \cite{Nandi_1997} as well as direct application to raw waveforms \cite{oshea_2016,clark_augmentation}.

To help understand how data quantity and quality affect the performance of the system there are three primary datasets used while training, and one unique evaluation set for evaluating the performance of all models that are trained.
The four datasets are described in Table \ref{tab:datasets}.
Captured data makes up the first dataset ($\Omega_C$), along with the evaluation set ($\Omega_{TC}$) such that the two sets are disjoint ($\Omega_C \cap \Omega_{TC} = \emptyset$).
The second set is a synthetic dataset ($\Omega_S$) that makes the waveforms in their pure form as shown in \eqref{eq:bb_sig} and adds synthetic errors associated with detection algorithms such as Frequency Offset (FO) and Sample Rate Mismatch (SRM), as well as varying the SNR to indicate different received power levels in the dataset.
The second and third dataset make use of a Joint Kernel Density Estimate (KDE) on $\Omega_C$ to mimic the distortions of FO, SRM, and SNR within them to attempt to minimize changing the distributions in the data to any extreme.
The third dataset, by contrast with $\Omega_S$, applies synthetic permutations to observations from $\Omega_C$ thereby creating an augmented dataset.

\tabdatasets
  
Additionally, in order to observe greater diversity in the application of dataset quality and quantity, the work shows the classification performance of three classification groups given in Table \ref{tab:waveforms}.

\tabwaveformsets


\subsection{RF Characteristics}
\label{sec:rf}


\noindent In the idealized world where the transceivers are in a physically stationary environment, RF signals can be thought of as processing signals at complex baseband (BB) with a channel between transmitter and receiver.
The transmitter's waveform is then represented as
\begin{equation}\label{eq:bb_sig}
    s_{BB}(t) = s_{\text{re}}(t) + j s_{\text{im}}(t),
\end{equation}
where the channel introduces a static set of unknowns: gain ($\alpha_0$); delay ($\tau_0$); and phase shift ($\theta_0$), as well as a time varying additive noise $\nu(t)$ to the receiver observation in addition to the transmitter's modulated baseband signal; however, with the received signal being modeled after perfect synchronization, eliminating the static unknowns with a perfect low pass filter results in the received signal given as
\begin{equation}\label{eq:ideal_bb_rx}
    r_{BB}(t) = s_{BB}(t) + \nu_{BB}(t).
\end{equation}
The ratio of power in the signal to that of the noise, or the Signal-to-Noise Ratio (SNR), often expressed in dB ($10\log_{10}(\int |s_{BB}(t)|^2dt/\int |\nu_{BB}(t)|^2dt)$), is then the primary limiting factor explaining the performance of the system, with $\nu_{BB}(t)$ most commonly assumed to be a circularly-symmetric complex Gaussian process.

\subsubsection{Real World Degradations}

\noindent In practice, the problem becomes vastly more complex as relative motion between transceivers, multiple transceivers, environmental noise, environmental motion, unintended radio emissions from manmade devices, mutlipath interference, and the imperfect hardware that transmits and receives the waveform are introduced.
An introduction to the effects of imperfect hardware is given by Fettweis \textit{et al} in \cite{dirty_rf} by looking at the individual degradations the hardware can add to a system as well as some mitigation strategies that can be applied; however, it is worth mentioning that these degradations are compounding and time varying, so while the worst of the effects can be calibrated out, the effects persist and cause a separation from the ideals assumed in \eqref{eq:ideal_bb_rx}.
For an example of how these degradations affect the ideal, here the frequency independent In-phase and Quadrature Imbalance (IQI) of the transceivers result in a carrier modulation functions that are ideally expressed as $\xi_{TX(ideal)}(t,f_c)=\exp(j2\pi f_c t)$ for the transmitter for a carrier frequency $f_c$ and $\xi_{RX(ideal)}(t,f_c)=\exp(-j2\pi f_c t)$ for the receiver as
\begin{equation}\label{eq:iq_imbal_mod}
\begin{split}
  \xi_{TX}(t,f_c) =& \bigg(\left(\frac{1+g_{TX}\exp(j\phi_{TX})}{2}\right)\exp(j2\pi f_c t)\\
  &+ \left(\frac{1-g_{TX}\exp(-j\phi_{TX})}{2}\right)\exp(-j2\pi f_c t)\bigg),\\
  \xi_{RX}(t,f_c) =& \bigg(\left(\frac{1+g_{RX}\exp(-j\phi_{RX})}{2}\right)\exp(-j2\pi f_c t)\\
  &+ \left(\frac{1-g_{RX}\exp(j\phi_{RX})}{2}\right)\exp(j2\pi f_c t)\bigg),
\end{split}
\end{equation}
where $g_X$ is the magnitude ratio imbalance, and $\phi_X$ is the phase difference between the quadrature mixer and the in-phase mixer \cite{dirty_rf}.
The ideal carrier modulators are recovered when the magnitude ratio imbalance is unity, $g_X=1$, and the phase difference is zero, $\phi_X=0$.
This results in a ideal received signal being changed from the ideal baseband transmitted waveform in \eqref{eq:ideal_bb_rx} into
\begin{equation}\label{eq:bb_rx}
\begin{split}
  r_{BB}(t) =& (s_{BB}(t)\xi_{TX}(t,f_c) + s_{BB}^*(t)\xi_{TX}^*(t,fc)\\
  &+ \nu_{BB}(t)e^{j2\pi f_c t} + \nu_{BB}^*(t)e^{-j2\pi f_c t})\\
  &\cdot \xi_{RX}(t,f_c) \ast h_{lp}(t)\\
  =& s_{BB}(t) + \text{IQI}(s_{BB}(t),g_{TX},g_{RX},\phi_{TX},\phi_{RX})\\
  &+ \nu_{BB}(t) + \text{IQI}(\nu_{BB}(t),1,g_{RX},0,\phi_{RX})
\end{split}
\end{equation}
where the $\text{IQI}(\cdot)$ is a function for the addition of IQI when both the transmitter's and receiver's parameters are known in the received signal as an additive interference given as
\begin{equation}
\begin{split}
  \text{IQI}(x(t),&g_{TX},g_{RX},\phi_{TX},\phi_{RX})=\\ 
  &x_{\text{re}}(t)\cdot(-j g_{RX}\sin(\phi_{RX}))\\
  &+ x_{\text{im}}(t)\cdot(-g_{TX}\sin(\phi_{TX}) \\
  &+ j(g_{TX}g_{RX}\cos(\phi_{TX}-\phi_{RX})-1)).
\end{split}
\end{equation}

\subsubsection{Understanding RF Data Origin}\label{sec:data_origin}
\noindent There are three common sources for data within ML dataset generation \cite{wong2020rfml}.
The first is the \textit{captured} or \textit{collected} data acquired by using a sensor and recording the data.
Under the most intuitive conditions, data collection performed using this approach in the application space provides the highest quality data to the problem to learn from because all unknown characteristics and sensor degradations will be present in the data \cite{clark_augmentation}.
However, when performing capture of rare events or while in search of other infrequent and uncontrollable events performing collection events can prove to be difficult to properly label, let alone find.
These problems along with having to procure and sustain the equipment and personnel to perform the collection often make collection in large quantities impractical and expensive.

\textit{Synthetic} datasets are therefore the most common and typically orders of magnitude cheaper to procure due to not being bound to waiting on real-world limitations.
For example, synthetic generation can occur in parallel for vastly different conditions with the limitation being computation resources, rather than the sensors and personnel in collection events.
The trade-off with synthesis is that significantly more information on the data is necessary in order to properly simulate, which without the appropriate knowledge can render useless models in the field \cite{clark_augmentation}.

The process of creating an \textit{augmented} dataset tries to bridge to strengths of captured and synthetic dataset creation, while covering their weaknesses \cite{wong2020rfml}.
By taking captured data and adding synthetic permutations of SNR, FO, and SRM the augmented dataset can smooth out missing observations from a limited collection event by having a better understanding of the detection characteristics of the sensors in use, while preserving all other real-world degradations native to the application space.

Here the different origins of data are kept separated from each other to get a better understanding of the characteristics of each approach, but the fusion of such datasets either by directly combining the datasets, or by performing staged learning should be done in practice.


\section{Materials and Methods}
\label{sec:data_qs}
\noindent The work performed in Clark \textit{et al} \cite{clark_augmentation} showed that, within the realm of AMC, the quantity of data has a functional relationship to the performance of a trained system given all other variables are constant.
Additionally, the work showed that the performance could be found as log-linear relation to the quantity of data for lower performance regions, but a log-sigmoidal relationship is more appropriate as performance reaches a maximum.
The process of regressing the relationship between quantity and performance was then suggested as a quantification measure of dataset quality in \cite{clark_data_quality}, where different datasets could then be compared across different quantities with the expected accuracy (e.g. dataset $A$ needs $X$ observations, while dataset $B$ needs $2X$ observations to achieve an accuracy of $90\%$), or other metric of performance, taken as the quality ($X|90\%$ or $2X|90\%$ in the previous example) of the dataset.
The inherent quality of any dataset can be described in three generalized terms of \textit{Comprehensiveness}, \textit{Correctness}, and \textit{Variety} \cite{ml_quality_1}.
In this work the datasets are already examined and confirmed to be \textit{Comprehensive} in that all the information being sought is included within the dataset, and \textit{Correct} in that the observations for each modulation are correctly identified and label.
The main concept of quality being examined is then that of \textit{Variety} or rather that the distributions on the observations within the datasets match, approximate, or deviate from the distributions of the test set, and therefore only the effect of quality in terms of \textit{Variety} can be examined in this work.

While these works gave an initial understanding of the data quantity and quality that fundamentally drive the process of an ML system, they provide minimal utility when trying to understand how much data is needed in order to achieve ideal performance and therefore reliably plan a data collection campaign.
For example, in Fig. 4 of \cite{clark_data_quality} looking at the 10-class classification performance, the log-linear fit predicts a performance of 90\% accuracy at roughly an order of magnitude less data than the corresponding log-sigmoidal fit, while both fits use the full range of trials available to regress the fit.
The results discussed above all depend on some initial \textit{good dataset} to contrast with, and while this work does not alleviate that requirement, here the question is answered of how to best use a limited \textit{good dataset} to forecast how much total data would be needed during training if neither the model, nor training approach is modified.

An ideal approach would be to use a metric that is both strongly correlated with the desired performance of the system, such as accuracy, in terms of Kendall's $\tau$ and has a relationship with data quantity that can be linearly derived from minimal data; however, a metric that reduces the error over that of performance directly regressed with quantity will be sufficient.
For this reason, the metrics that have been developed to predict the transferability of a pre-trained model onto a new \textit{target} dataset, discussed in Section \ref{sec:tl}, are repurposed to predict data quantity requirements and provide a new metric of quality for a model's training dataset with regard to the \textit{target} dataset, which is the evaluation dataset in this work as shown in Fig. \ref{fig:dataviz} (c).

\subsection{Examining Correlation between Performance and Metrics}
\noindent The first step is confirmation that the chosen metrics correlate in a beneficial manner with the performance value of interest, classification accuracy in this case.
In order to understand whether a metric is well correlated with classification accuracy, the weighted Kendall's $\tau$ is calculated using the SciPy implementation \cite{scipy} and found for three datasets (Table \ref{tab:datasets}: $\Omega_{C},\Omega_{A},\Omega_{S}$) and compared against three sets of modulation classification sets (Table \ref{tab:waveforms}: $\Phi_{3},\Phi_{5},\Phi_{10}$).
The Kendall's $\tau$ weighted correlations are presented in Table \ref{tab:wkt} and show high values of correlation for all three metrics in the case of $\Omega_{C}$ and $\Omega_{A}$ datasets; however, the correlation for the $\Omega_{S}$ dataset shows a worse correlation between accuracy and all three metrics.
Looking at the performance at the relationships between performance and the proposed metrics in Fig. \ref{fig:acc_metric_relationship} shows that the performance and metrics are tightly clustered, while for $\Omega_{C}$ and $\Omega_{A}$, definite trends are observable.
Looking at the performance of the different datasets as a function of quantity used during training in Fig. \ref{fig:qty_acc_relationship} helps to further explain this decrease in correlation in that the performance results of networks trained on $\Omega_{S}$ are comparably independent from the quantity of data used for the synthetic observations.
Therefore, the classification accuracy and metrics extracted from the networks trained on $\Omega_{S}$ are more akin to noisy point measurements rather than a discernable trend to examine.

\tabkendallstau

\figAMrelationship
\figQArelationship

The main observation is that when there is a discernable trend between performance and data quantity, the correlation of all three metrics are considerably high, and therefore are potential metrics with which to regress the relationship with data quantity in search of a quantity estimator for the total data needed to achieve a desired performance.

\subsection{Regression of Quantity and Metrics}
\noindent With the confidence that the TL metrics discussed above have a positive and significant correlation with the performance of the system when performance increases with regard to the quantity of data used during training, the goal is to now derive the relationship between those metrics and data quantity, with preference being given to the metric that has a better goodness-of-fit (GoF) with a form of linear regression.
In this case, a log-linear regression is used between the metrics and the data quantity.
Starting with the accuracy of each network as shown in Fig. \ref{fig:qty_acc_relationship}, the log-linear fit is able to provide a quality value in terms of the accuracy achievable for a given observations per class (OPC) for the three datasets.
Looking at the $\Phi_{10}$ problem set shows the quality quantification as
\begin{itemize}
    \item $\Omega_C \rightarrow 81\%$ accuracy $|\; 1$M OPC                  
    \item $\Omega_S \rightarrow 20\%$ accuracy $|\; 1$M OPC
    \item $\Omega_A \rightarrow 76\%$ accuracy $|\; 1$M OPC,
\end{itemize}
but the quality can just as easily be defined as the OPC needed in order to achieve a given accuracy given the linear fit can be inverted as
\begin{itemize}
    \item $\Omega_C \rightarrow 5.25$M OPC $|\; 90\%$ accuracy
    \item $\Omega_S \rightarrow \infty$ OPC $|\; 90\%$ accuracy
    \item $\Omega_A \rightarrow 7.04$M OPC $|\; 90\%$ accuracy.
\end{itemize}
However, the log-linear regression between data quantity and accuracy has an undesired effect between the data points and the linear fit, which is that at the ends of the available data there is increased error relative to the center of the data points.
Additionally, because the sign of error is the same at both ends, this suggests that the linear fit between data quantity and accuracy when there is minimal data will severely underestimate the data quantity needed to achieve high performance systems.
For a better look at this issue, Fig. \ref{fig:qty_acc_residual} examines the residuals for the $\Phi_{10}$ waveform set across the three dataset types.

\figPTenAresiduals

This same sign of error at the ends of the available data suggest that a non-log-linear fit would be more appropriate for regressing the relationship between accuracy and data quantity, which is poorly suited to understanding the full relationship as available data becomes more limited to a narrow subset of the full data range.
For example just looking at a narrow portion of either end, high or low data quantity, does not provide enough context to predict a good non-linear fit.
Therefore, a GoF measure that weights the outer errors more significantly than the errors toward the center of the data range is desired.
Additionally, since both edges of the residual are of equal significance and the results are non-uniformly sampled across the observation space, a weighting that balances the weights into histogram bins will be used to normalize equal significance in the edges of the GoF measure.
For simplicity, three bins will be used indicating lower, mid, and high data quantity observations relative to the log-linear fit.
The weights are suggested as 
\begin{equation}
\begin{split}
    w'(q_x[i]) &= 
\begin{cases}
    \left|q_x\right|/\left|b_{low}\right| & q_x[i] \in b_{low} \\ 
    \left|q_x\right|/3\left|b_{mid}\right| & q_x[i] \in b_{mid} \\ 
    \left|q_x\right|/\left|b_{hi}\right| & q_x[i] \in b_{hi}
\end{cases} \\
    w(q_x[i]) &= \frac{w'(q_x[i])}{\sum_{j=1}^{|q_x|} w'(q_x[j])}
\end{split}
\end{equation}
where the middle bin has one third the weight of the edges, which without, would have all three regions equally weighted.
The division edges between bins is taken as evenly spaced on log scale between the minimum and maximum data quantities in the set, with $|q_x|$ being the number of elements in the set, while $\{|b_{low}|,|b_{mid}|,|b_{hiw}|\}$ are the number of observations within that bin.
Those weights are then normalized such that their sum is unity.
The GoF is then taken as the Normalized Root Weighted Mean Squared Error (NRWMSE)
\begin{equation}
    \text{gof}(\alpha_x,q_x,\hat{f}_{ll}) = \sqrt{\frac{\sum_{i=1}^{|q_x[i]|} w(q_x[i])\cdot(\hat{f}_{ll}(q_x[i]) - \alpha_x[i])^2}{\text{Var}(\alpha_x)}}
\end{equation}
where the quantities ($q_x$) and accuracies ($\alpha_x$) are use to derive the log-linear fit ($\hat{f}_{ll}$); however, the accuracies and fit can be swapped out for any other metric and matching fit.

The GoF for accuracy, NCE, LEEP, and LogME metrics are given in Table \ref{tab:gof}.
A general conclusion is that all three metrics have potential to provide a better prediction of data quantity needed to achieve high performance; however, considering the correlation presented in Table \ref{tab:wkt} in addition to these results suggest that NCE will be the most consistent estimate, with LEEP being a close second.
LogME, by comparison, offers the most promise with regard to the augmented dataset, but has the highest variability among the three metrics examined here.
One more unique attribute about the linear regressions of the metrics, is that accuracy, NCE, and LEEP all have residuals typically indicating that the true quantity of data that is needed will be underestimated, while LogME's residuals are inverted suggesting that that LogME's regression will overestimate the amount of data, giving soft bounds of the required quantity of data being between the estimates of NCE and LogME predictions.
\tabgof

\subsection{Predicting The Data Quantity Needed}
\noindent Now that the metrics have been compared in terms of a regressed log-linear fit with the quantity of data used to train the model, the question is how to determine what value of the metrics will provide a desired performance.
Looking back at Fig. \ref{fig:acc_metric_relationship} shows that the metrics and accuracy don't have an easily fit relationship that would map a metric back to accuracy, and in fact would only be trading one non-linear regression for another.
To overcome this problem, label whitening to acquire near perfect performance is proposed to act as a quasar that can help map the performance of the metrics with accuracy.

The procedure starts with label smoothing \cite{smoothing} \eqref{eq:label_smooth} of the truth labels for the evaluation set, followed by a logit transform \eqref{eq:logit}, which without the label smoothing would not be a useful approach as infinite values would be returned for the correct class and negative infinity for all other classes.
\begin{equation}\label{eq:label_smooth}
    \tilde{l}_x = l_x - \gamma\cdot\left(l_x-C^{-1}\right)
\end{equation}
\begin{equation}\label{eq:logit}
    m_x = \log\left(\frac{\tilde{l}_x}{1-\tilde{l}_x}\right)
\end{equation}
Label smoothing applied on its own does not affect the value of accuracy, NCE, nor LogME, but it does affect LEEP score and is dependent on the smoothing factor, $\gamma$, and number of classes in the classification problem, $C$.
The effect of $\gamma$ on the LEEP metric can significantly affect the metric, so $\gamma$ is chosen to be the minimum value that the approach of $|l_x-\tilde{l}_x|>0$ within the chosen machine precision.
The effect of label smoothing and the logit transform allows for the values to now sit at a finite coordinate to which noise can be added to stochastically decrease the accuracy of the system in a controlled manner.
The normal distribution is used to whiten the logits in this case where the standard deviation of the noise, $\sigma$, can be chosen for a degradation of accuracy, $\epsilon$, of the true labels given the number of classes in the problem space and the label smoothing $\gamma$ in use.
\begin{equation}
    \tilde{m}_x = m_x + \mathcal{N}(0,\sigma^2)
\end{equation}
\begin{equation}
    \sigma(\epsilon) = \frac{\log\left(C^2(1-\gamma) + \gamma^2(C-1)\right) - \log\left(\gamma^2(C-1)\right)}{2\cdot\text{erf}^{-1}\left(2\cdot\sqrt[C-1]{1-\epsilon}-1\right)}
\end{equation}
With the whitened logits the inverse logit, or logisitic, transform is applied and balanced such that the sum of any result is unity, $\sum\hat{l}_{xi} = 1\ \forall\ i$.
\begin{equation}
    \hat{l}_x = \frac{\exp(\tilde{m}_x)/(1+\exp(\tilde{m}_x))}{\sum_{k\in C}\exp(\tilde{m}_x[k])/(1+\exp(\tilde{m}_x[k]))}
\end{equation}
Fig. \ref{fig:whitening} shows the effects of this procedure on the error and metrics for a given $\epsilon$ averaged over 1000 iterations, and shows a trend that can be maintained with increasing $\epsilon$; however, an important note is that this type of error does not properly reflect the distributions of error that can be expected, so smaller values ($\leq 1e-5$) of $\epsilon$ will likely be more appropriate than larger values ($0.1$).
Looking at the residual error, \eqref{eq:resid_err}, in terms of the dependent variable, $\epsilon$, relative to the measured value, $\hat{\epsilon}$, as seen in the top left plot of Fig. \ref{fig:whitening}, the minimum average error across the three classes is achieved at $\Delta(1e\text{-}5,\bar{\hat{\epsilon}})=0.0169$, with the average normalized residuals being nearly equal at the extremes ($\Delta(1e\text{-}8,\bar{\hat{\epsilon}})=0.178;\ \Delta(1e\text{-}1,\bar{\hat{\epsilon}})=0.182$).
\begin{equation}\label{eq:resid_err}
    \Delta(\epsilon,\hat{\epsilon}) = \frac{\epsilon - \hat{\epsilon}}{\epsilon}
\end{equation}

\figWhitening

At this point a means for determining the value for each metric has been proposed that won't suffer from the need to have a perfect response that can be used, and will help with metrics such as LogME where the maximum is not immediately known given the iterative solution that is employed to produce the score.
These values for a given small $\epsilon$ can then be used to regress the corresponding metric's data estimate for achieving such performance.
The log-linear regressions for each metric, dataset, and waveform space combinations are shown in Fig. \ref{fig:qty_metric_relationship}, while the log-linear regressions for accuracy are shown in Fig. \ref{fig:acc_metric_relationship}, and together help to better visualize the GoF results given in Table \ref{tab:gof}.

\figMetricRelationships

Making use of the whitening procedure above and the log-linear regressions between data quantity and the metric's score a prediction for data quantity needed to achieve arbitrarily high performance can then be found.
For example applying an error of $\epsilon = 1e\text{-}5$ to each examined problem space for the metrics and selecting a value averaged over 1000 iterations, the data quantity predictions can be made for each metric as shown in Table \ref{tab:perf}.
Where the predictions are found by inverting the linear fit to estimate the quantity from the predicted metric as
\begin{equation}
    \tilde{q}_{\chi} = \log_{10}(\hat{q}_{\chi}) = \frac{\hat{M}_{\chi} - b_{x}}{s_{\chi}},
\end{equation}
where the $\tilde{q}_{\chi}$ value is the logarithm base ten of the quantity estimate for the selected metric $\chi \in [$Accuracy, NCE, LEEP, LogME$]$, $\hat{M}_{\chi}$ is the metric value found through the whitening procedure, and $s_{\chi},\ b_{\chi}$ are the slope and y-intercept, respectively, of the log-linear regression given a logarithm base ten applied.
\tabperf


\section{Results and Discussion}
\label{sec:application}
\noindent The prior sections made use of all data points taken in order to establish the best predictions for data quantity with their given metric.
As these are estimates that are intended to predict the data quantity needed to achieve high performance systems through the increase of available data alone, certifying any result in particular is beyond the scope of this work, as the expected predicted quantities will far exceed the available data acquired.
Instead, the focus shifts to how less available data during training relatively affects the prediction capability for each metric in comparison to greater quantities of available data.

Due to the performance of the synthetic dataset stagnating, further analysis will ignore this case going forward.
For the purpose of finding how well the log-linear regression with each metric is able to predict the data quantity needed, the data quantities provided in Table \ref{tab:perf} with preference for a quantity estimate given by the GoF in Table \ref{tab:gof} will be used such that the metric that achieved the best GoF will be used as the truth for the problem space.
Therefore predictions of the models making use of $\Omega_C$ will use the LEEP metric's quantity prediction for $\Phi_3$ and $\Phi_5$, but will make use of the NCE prediction for $\Phi_{10}$, while the predictions for $\Omega_A$ will all make use of the LogME metric and these quantity prediction are summarized in Table \ref{tab:limited}.
\tablimited
The predictions for each metric can be seen in Fig. \ref{fig:qty_predictions} where the top row shows the predictions when using the $\Omega_C$ dataset, while the bottom shows the predictions for the $\Omega_A$ dataset.
The columns consist of waveform spaces \{$\Phi_{3}$,$\Phi_{5}$,$\Phi_{10}$\} from left to right respectively.
\figMetricPredictions

The general understanding given in Fig. \ref{fig:qty_predictions} is that both NCE and LEEP will give a more realistic prediction for data quantity than Accuracy alone, while the prediction given by LogME can serve as an upper bound.
Due to the log-linear regression any deviation can result in orders of magnitude error in either underestimation or overestimation, and since without having enough data to acquire the metric that produces the best GoF regression, a midpoint estimate is recommended with only minimal data available.
The midpoint estimate seeks to balance the two extremes such that the estimate becomes $\tilde{q}_{\text{MidPoint}} = 
0.5\cdot(\tilde{q}_{\text{NCE}}+\tilde{q}_{\text{LogME}})$, such that the midpoint estimate averages the quantity estimates on the log scale rather than the linear.

Returning to Table \ref{tab:limited}, an important note of worth is understanding how long a sequential collection of data of this kind would take in order to accomplish, that is for a collection that records at 10kHz, these waveforms from three waveform groups \{$\Phi_{3}$,$\Phi_{5}$,$\Phi_{10}$\}, collection of the number of observations implied would require [1.72, 144.5, 33.4] years to acquire for the target observations needed for the $\Omega_C$ predictions and require [0.989, 82.9, 19.1] terabytes of storage to store in an uncompressed state.
While this could be feasible if the collection was performed in parallel rather than a sequential collection, the suggestion that should be taken rather than immediately starting a long term collection is to instead improve the training routine and model architecture to instead allow for this procedure to produce a regression with a more significant slope than the approach used to produce these results.

\section{Conclusion}
\label{sec:conclusion}
\noindent The problem of estimating the amount of data needed in order to achieve a high performing ML model for AMC problems is discussed within this work.
While if large quantities of data are already available, $>1e9$ observations per class in the problem presented in this work, a log scale nonlinear regression between performance and data quantity can help to determine how much more data is needed, this is often not feasible for problems with only a small amount of data on hand (i.e. $<1e5$ observations per class) due to the potential for the asymptotic bends in performance not being visible leading to the performance regression significantly underestimating the amount required.
Making use of the metrics developed for TL in order to chose a model to help linearize, on a log scale, the relation between a metric and data quantity increased accuracy predictions can be made by utilizing the metrics NCE, LEEP, and LogME.
These metrics in turn can help bound to amount of data that would be needed with a current approach, and help determine whether a large scale data collection should take place or if further refinement to the training procedure and model architecture are a better approach given the program's constraints.
Given the tendency for NCE and LEEP metrics to underestimate the amount of data needed, and the tendency of LogME to overestimate the amount of data a midpoint approach is proposed, on a log scale, between NCE and LogME to offer a balanced estimate.
Given that NCE and LEEP offer similar performance estimates and GoF measures on the collected dataset problems, and LogME is seen to offer a better GoF for the augmented dataset problems balancing the two offers a more reasonable measure under data constrained conditions when performing the estimate.

While this approach shows traction within the RFML problem space of AMC, additional research is still needed in order to understand if these techniques can be more widely applied to other classification problem spaces in ML in general.

\begin{adjustwidth}{-\extralength}{0cm}

\reftitle{References}

\PublishersNote{}
\end{adjustwidth}
\end{document}